%% file: eacl2023.tex
\pdfoutput=1

\documentclass[11pt]{article}

\usepackage{EACL2023}

\usepackage{times,latexsym}
\usepackage{url}
\usepackage[T1]{fontenc}
\usepackage{graphicx}
\usepackage{times}
\usepackage{latexsym}
\usepackage{amsmath}
\usepackage{longtable}
\usepackage{booktabs}
\usepackage[utf8]{inputenc}
\usepackage[T1]{fontenc}
\usepackage[scaled=0.8]{beramono}
\usepackage{microtype}
\usepackage[normalem]{ulem}
\usepackage{enumitem}
\usepackage{inconsolata}

\usepackage{tikz}
\usetikzlibrary{calc}
\usetikzlibrary{decorations.pathmorphing}
\usetikzlibrary{shapes,arrows,chains}
\usetikzlibrary{shapes.geometric}
\usetikzlibrary{patterns}
\usetikzlibrary{positioning}
\tikzset{>=latex}
\usepackage{pgfplots}
\usepackage{relsize}

\title{Noisy Parallel Data Alignment}

\author{Ruoyu Xie, Antonios Anastasopoulos \\
  Department of Computer Science, George Mason University \\
  \texttt{\{rxie, antonis\}@gmu.edu} }

\begin{document}
\maketitle
\begin{abstract}
An ongoing challenge in current natural language processing is how its major advancements tend to disproportionately favor
resource-rich languages, leaving a significant number of under-resourced languages behind. Due to the lack of resources required to train and evaluate models, most modern language technologies are either nonexistent or unreliable to process endangered, local, and non-standardized languages. Optical character recognition (OCR) is often used to convert endangered language documents into machine-readable data. However, such OCR output is typically noisy, and most word alignment models are not built to work under such noisy conditions. In this work, we study the existing word-level alignment models under noisy settings and aim to make them more robust to noisy data. Our noise simulation and structural biasing method, tested on multiple language pairs, manages to reduce the alignment error rate on a state-of-the-art neural-based alignment model up to 59.6\%.\footnote{Data and code are available online: \url{https://github.com/ruoyuxie/noisy_parallel_data_alignment}}
\end{abstract}

\section{Introduction}
Modern optical character recognition (OCR) software achieves good performance on documents in high-resource standardized languages, producing machine-readable text which can be used for many downstream natural language processing (NLP) tasks and various applications~\cite{ignat2022ocr,van2020assessing,amrhein2018supervised}. However, attaining the same level of quality for texts in less-resourced local and non-standardized languages remains an open problem~\cite{rijhwani-etal-2020-ocr}. 

The promise of OCR is particularly appealing for endangered languages, for which material might exist in non-machine-readable formats, such as physical books or educational materials. Digitizing such material could lead to the creation of NLP technologies for such otherwise severely under-served communities~\cite{bustamante2020no}.

Beyond the primary goal of digitizing printed material in endangered languages, the need for robust alignment tools is wider. The majority of the world's languages are being traditionally oral~\cite{bird-2020-sparse}, which implies that to obtain textual data at scale one would need to rely on automatic speech recognition (ASR), which in turn would produce invariably noisy outputs. It is worth noting that the availability of translations can significantly improve systems beyond machine translation (MT), such as OCR~\cite{rijhwani-etal-2020-ocr} or ASR~\cite{anastasopoulos2018leveraging}. This creates a chicken-and-egg situation: on one hand, OCR and ASR can be used to obtain noisy parallel data; on the other hand, having good quality aligned data can improve OCR or ASR.

\begin{figure}[t]
    \centering  
    \vspace{-1em}
    \includegraphics[width=7.5cm]{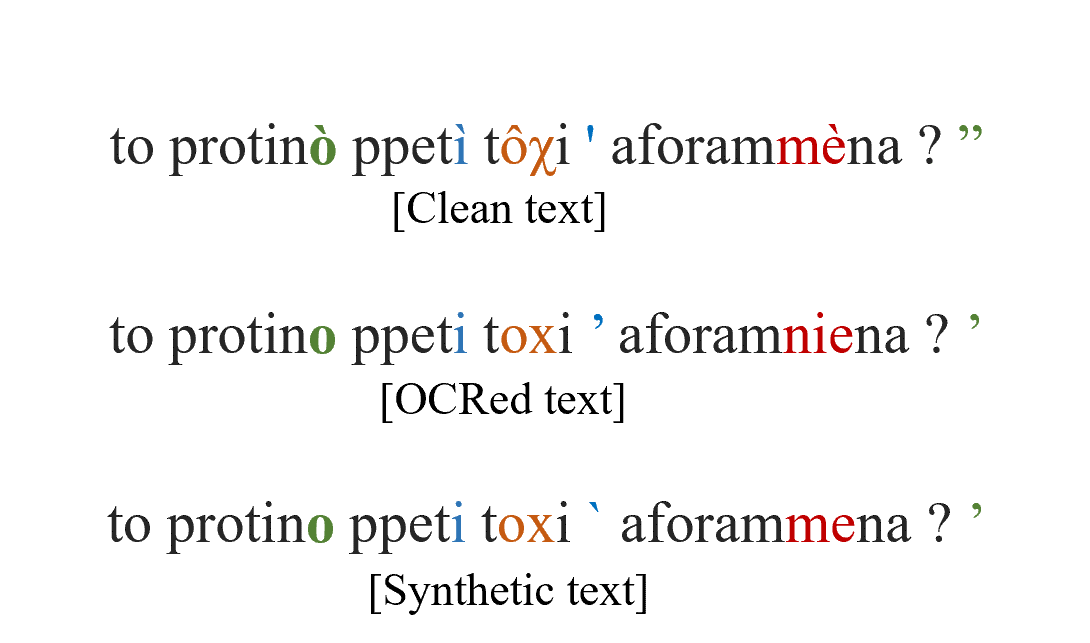}
    \caption{Synthetic data example in Griko with character differences highlighted. Our synthetic data manage to mimic the real OCR noise. 
    }
    \label{fig:example}
    \vspace{-1em}
\end{figure}

In this vein, We focus on the scenario of digitizing texts in a less-resourced language along with their translations (usually high-resource and/or widely spoken) similar to~\citet{rijhwani-etal-2020-ocr}. 
Digitizing parallel documents can also be beneficial for educational purposes, as one could then create dictionaries through word- and phrase-level alignments, or ground language learning on another language (a learner's either L1 or L2).
Also, as~\citet{ignat2022ocr} showed in recent work, such parallel corpora can be meaningfully used to create MT systems. However, the process that transforms digitized books or dictionaries into parallel sentences for training MT systems requires painstaking manual intervention.

In theory, the process could be semi-automated using sentence alignment methods, but in practice, the situation is very different: OCR systems tend to generate very noisy text for endangered languages~\cite[][\textit{inter alia}]{alpert2016machine}, which in turn leads to poor alignments between two parallel sides. As we show, alignment tools are particularly brittle in the presence of noise.

In this work, we take the first step towards solving the above issue. We investigate the relationship between OCR noise and alignment results and build a probabilistic model to simulate OCR errors and create realistic OCR-like synthetic data. We also manually annotate a total of 4,101 gold alignments for an endangered language pair, Griko-Italian, in order to evaluate our methods in a real-world setting. We leverage structural knowledge and augmented data, greatly reducing the alignment error rate (AER) for all four high- and low-resource language pairs up to 59.6\%.

\section{Problem Setting}
\label{sec:problem}

Our work is a straightforward extension of previous word-level alignment work.
Given a sequence of words $\mathbf{x}=(x_{1},\dots,x_{n})$ in a source language and $\mathbf{y} = (y_{1},\dots,y_{m})$ in a target language, the alignment model produces alignment pairs:\footnote{Sometimes denoted with a latent variable, but we use an equivalent notation for simplicity.} 
\vspace{-.5em}
\begin{equation*}
   \mathcal{A} = \{(x_{i},y_{j}): x_{i}\in \mathbf{x},y_{j} \in \mathbf{y}\}\vspace{-.5em}
\end{equation*}
The difference with previous work is that the starting data will be the output of an OCR pipeline, hence producing noisy parallel data $(\mathbf{x^*},\mathbf{y^*})$ instead of ``clean''  $(\mathbf{x},\mathbf{y})$ ones. The level of noise may vary between the two sides.

Hence, our goal is to produce an alignment
\vspace{-.5em}
\begin{equation*}
   \mathcal{A}^* = \{(x_{i},y_{j}): x_{i}\in \mathbf{x^*},y_{j} \in \mathbf{y^*}\}\vspace{-.5em}
\end{equation*}
that will be as close to the alignment $\mathcal{A}$ that we would have obtained without the presence of noise. We measure model performance using the alignment error rate~\cite[AER;][]{och2003systematic} against the gold alignments.\footnote{Lower AER means a better alignment. More details on the metric in Appendix~\ref{metric}.}

\section{Method}
We create synthetic data that mimic OCR-like noise, that can be used to train/finetune alignment models. Our \textit{\textbf{simple yet effective}} method mainly consists of \textit{(i)} building probabilistic models based on edit distance measures and capturing real OCR errors; \textit{(ii)} creating synthetic (noisy) OCR-like data by applying our error-introducing model on clean parallel data; \textit{(iii)} training or finetuning alignment models on synthetic data.

\subsection{OCR Error Modeling}
\paragraph{Error types} For OCRed text, different types of texts, languages, and corpora will lead to different error distributions. At the character level, there are generally three types of OCR errors: insertions, deletions, and substitutions. In most cases, deletions and substitutions are more common, with spurious insertions being rarer. 

\paragraph{Noise model} By comparing the OCRed text with its post-corrected version, we use Levenshtein distance to compute the edit distances and the probability distributions of edits/errors over the corpus with a straightforward count-based approach.

We treat deletion error as part of substitution error. Given a sequence of characters \(x_{i},\dots,x_{j}\) from a clean corpus $\mathbf{x}$ and a sequence of characters \(y_{i},\dots,y_{j}\) from its OCRed noisy version $\mathbf{y}$, we simply count the number of times a correct character $x_i$ is recognized as character $y_i$ (or recognized as the empty character $\epsilon$ if it is erroneously deleted). We can then compute the probability of an erroneous substitution or deletion as follows:

\[
   P_{sub}(x_{i}\rightarrow y_{i}) = \frac{\text{count}(x_{i}\rightarrow y_{i})}{\text{count}(x_{i})}
\] 

and the overall substitution error distribution is conditioned on the correct character $x_i$:

\begin{equation*}
   \mathcal{D}_{sub}(x_i) \sim P_{sub}(x_{i}\rightarrow y_{i}).
\end{equation*}

For insertion errors, we consider that insertion occurs when $\epsilon$ becomes another character $y_i$ and count the number of times that insertion occurs after its previous character. A special token \texttt{<begin>} is used when insertion occurs at the beginning of the sentence. In general, we calculate the insertion error probability with: 
\[
   P_{ins}(x_{i-1}\epsilon\rightarrow x_{i-1}y_i)\!=\!\frac{\text{count}(x_{i-1}\epsilon\rightarrow x_{i-1}y_i)}{\text{count}(x_{i-1}\epsilon)}
\]
and the insertion error distribution for $x_{i-1}$:
\begin{equation*}
   \mathcal{D}_{ins}(x_{i-1}\epsilon) \sim P_{ins}(x_{i-1}\epsilon\rightarrow x_{i-1}y_i).
\end{equation*}

\subsection{Data Augmentation}
Synthetically noised data can be created by leveraging the calculated probability distributions from the previous section and traversing through the clean corpus for every character. 

For each character $\mathbf{c}$, we obtain its probability to be erroneous in the OCR output by sampling from the distribution of the substitution and insertion probabilities $\mathcal{D}_{ins}(c),\mathcal{D}_{sub}(c)$.\footnote{Including a third option for not inserting an error} We randomly decide whether to add an error here based on its error distribution.

If an error will be introduced on $\mathbf{c}$, we then randomly choose its corresponding error based on \(P_{sub}(c)\) or \(P_{ins}(c)\) depending on either substitution or insertion operation receptively. 

Our method attempts to mimic the real OCR errors in given languages and corpus, resulting in very similar noise distributions. Figure~\ref{fig:example} shows a side-by-side comparison of three versions of the same sentence, to showcase how realistic our synthetic text is.

\subsection{Model Improvement}

Given our synthetically noised parallel data, and potentially along with the original clean parallel data, we can now train or finetune a word alignment model to improve the model performance. 

In the case of unsupervised models like the IBM translation models~\cite{brown1993mathematics}, \texttt{fast-align}~\cite{dyer2013simple}, or \texttt{Giza++}~\cite{och2003minimum}, we simply train on the concatenation of all available data. 

We also work with the state-of-the-art neural alignment model of ~\citet{dou-neubig-2021-word}, which is based on Multilingual BERT~\texttt{(mBERT)}~\cite{devlin2018bert}.\footnote{See Section~\ref{subsec:setup} for more details.} For this model, we distinguish two cases: supervised and unsupervised finetuning.\footnote{We use provided default parameters for both cases, which can be found on https://github.com/neulab/awesome-align} Under a supervised setting, we first obtain silver alignments from the clean dataset and use them as targets for the synthetic noisy data. The unsupervised setting is conceptually similar to training models like \texttt{Giza++}: we feed synthetically-noised sentence pairs into the alignment model, without using the target alignment as supervision. In low-resource scenarios, we leverage a diagonal bias to further improve the model's performance. 
\label{sec:modelImprovement}

\section{Languages and Datasets}
We study four language pairs with varying amounts of data availability: English-French, English-German, Griko-Italian, and Ainu-Japanese.\footnote{Griko and Ainu are both under-resourced endangered languages.}

\subsection{Dataset for Error Extraction} 
The ICDAR 2019 Competition on Post-OCR Text Correction~\cite{rigaud2019icdar} dataset provides both clean and OCRed text for English-French and English-German, which we use our noisy model to learn and mimic OCR errors for English, French, and German. 
\label{extract}

For Griko-Italian and Ainu-Japanese, \citet{rijhwani-etal-2020-ocr} provide around 800 OCRed noisy and clean (post-corrected) sentences for both Griko and Ainu, from which we extract error distributions; for Italian and Japanese, only OCRed text is provided.\footnote{The quality of the OCR model on these high-resource languages are generally reliable.} 

To understand the characteristics of our datasets, we report the observed CER in Table~\ref{table:CERedit}. Generally, substitutions are the most common errors. Notice that Griko and Ainu have seemingly lower scores than any high-resource languages; that's because both use the Latin alphabet, the data that were digitized are typed in books with high-quality scans.\footnote{The English, French, and German data from ICDAR have lower-quality scans.}

\begin{table}[t]
\centering
\begin{tabular}{c c c c} 
 \toprule
 Language& Total CER & Sub. \% & Ins. \% \\
 \midrule
English& 7.4 & 79 & 21\\ 
German& 4.9 & 87 & 13\\ 
French& 4.8 & 85.7 & 14.3\\ 
Griko& 3.3 & 96.8 & 3.2\\ 
Ainu& 1.4 & 91.9 & 8.1\\ 

\bottomrule

\end{tabular}
\caption{Total character error rate (CER) and percentage of substitution and insertion errors. Generally, substitution is the most common error in OCR output.}
\label{table:CERedit}
\vspace{-1em}
\end{table}

\begin{table}[t]
\centering
\begin{tabular}{c c c c} 
 \toprule
 Language& Real CER & Syn. CER  & Diff.\\
 \midrule
English& 7.4 & 6.5 & 0.9\\ 
German& 4.9 & 6.7 & 1.8\\ 
French& 4.8 & 5.3 & 0.5\\ 
Griko& 3.3 & 3.3 & 0\\ 
Ainu& 1.4 & 1.0 & 0.4\\ 

\bottomrule

\end{tabular}
\caption{Our synthetically-noisy data have similar CER compared to the real OCR outputs, which implies that the real OCRed noisy data can be mimicked by our noise simulation model.}
\label{table:CERCompare}
\vspace{-1em}
\end{table}

\subsection{Synthetic Data} 
We create synthetic data by applying captured OCR noise on clean text. For English, French, and German, the clean text comes from Europarl v8 corpus~\cite{koehn2005europarl}. For Ainu, there are 816 clean sentences from \citet{rijhwani-etal-2020-ocr}, from which we keep the first 300 lines as test set and use the rest to create synthetic data. ~\citet{anastasopoulos+al:2018:COLING} provide 10,009 clean sentences for Griko.
Table~\ref{table:CERCompare} shows the CER comparison between our synthetic data and real OCR data. 

\subsection{Test Set and Gold Alignment}
The test set and gold alignment for English-French come from~\citet{mihalcea2003evaluation}. For English-German, the test set and gold alignments come from Europarl v7 corpus~\cite{koehn2005europarl} and~\citet{vilar2006aer}, respectively. 
To study the effect of OCR-like errors on alignment, we create synthetically-noised test sets for both languages pairs by applying noise on one side or both, which results in four copies of the same test set: clean-clean, clean-noisy, noisy-clean, and noisy-noisy.~\label{4testset}

For low-resource language pairs,~\citet{rijhwani-etal-2020-ocr} provide about 800 parallel sentence pairs for each. We use the first 300 sentence pairs as our test sets. For the purpose of fair evaluation in our method, we annotate a total of 4,101 \textit{gold} word-level alignment pairs for Griko-Italian test set. 
On the other hand, we obtain \textit{silver} alignments from \texttt{awesome-align} for Ainu-Japanese as there is no existing gold alignment data available.\footnote{While not ideal, we can still measure how different results are when comparing alignments on clean versus noisy data.}

\section{Experiments}
In this section, we present multiple experiments and demonstrate that our method results in significant AER reductions.

\subsection{Experimental Setup}
\label{subsec:setup}

\paragraph{Models} We study the following models:

\begin{itemize}[leftmargin=*,noitemsep,nolistsep]
\item \texttt{IBM model 1\&2}~\cite{brown1993mathematics}: the classic statistical word alignment models. They underpinned many other statistical machine translation and word alignment models.
\item \texttt{Giza++}~\cite{och2003minimum}: a popular statistical alignment model that is based on a pipeline of IBM and Hidden
Markov models~\cite{vogel1996hmm}.
\item \texttt{fast-align}~\cite{dyer2013simple}: a simple but effective statistical word alignment model that is based on IBM Model 2, with an additional bias towards monotone alignment.
\item \texttt{awesome-align}~\cite{dou-neubig-2021-word}: a neural word alignment model based on \texttt{mBERT}. It finetunes a pre-trained multilingual language model with parallel text and extracts the alignments from the resulting representations.
\end{itemize}

\begin{table}[t]
\centering
\begin{tabular}{rccc} 
 \toprule
Model & Clean & OCRed & Diff.  \\
\midrule
\texttt{IBM 1}& 43.7 & 49.2& 5.5\\ 
\texttt{IBM 2}& 37.3 & 43.4 & 6.1\\ 
\texttt{Giza++}& \textbf{14.5} & \textbf{20.8} & 6.3\\ 
\texttt{fast-align}& 19.8 & 25.7 & 5.9\\ 
\texttt{awesome-align}& 45.1 & 48.8 & \textbf{3.7}\\ 
\bottomrule

\end{tabular}
\caption{AER comparison for Griko-Italian. \texttt{Giza++} performs best on both settings, but it exhibits the largest drop in performance.} 
\label{table:noisyEffect}
\end{table}

\begin{table*}[t]
\centering
\begin{tabular}{@{}l|cc|cc} 
 \toprule
 & \multicolumn{2}{c|}{Griko-Italian}& \multicolumn{2}{c}{Ainu-Japanese} \\
& Clean & OCRed &  Clean & OCRed  \\
\midrule
\textsc{base}& 45.1 & 48.8  & 28.2 & 29.2  \\ 
\textsc{unsup-ft} (\texttt{A}) & 23.2$\pm$1.6 & 28$\pm$1.1 & 21.1 $\pm$1 & 22.2$\pm$1.2 \\ 
\textsc{sup-ft} (\texttt{B}) & 22.3$\pm$2 & 26.6$\pm$1.1 & 30.9$\pm$2.1 & 31.5$\pm$1.2 \\ 

\midrule
\multicolumn{5}{l}{+ structural bias} \\

\textsc{unsup-ft} (\texttt{A}) & 18.7$\pm$1 & 24.2$\pm$0.6 &  \textbf{15.3$\pm$3.9} & \textbf{13.8$\pm$4.2}  \\ 
\textsc{sup-ft} (\texttt{B}) & \textbf{18.2$\pm$2.6} & \textbf{22.9$\pm$2.4} &  26.3 $\pm$2.1 & 27.4$\pm$1.8  \\ 

\midrule
AER reduction&59.6\% & 53.1\% & 45.7\% & 52.7\%\\

\bottomrule
\end{tabular}
\caption{For both endangered languages, our approach greatly reduces AER for both clean and OCRed data.} 
\label{table:proposedModels}
\vspace{-1em}
\end{table*}

\subsection{The Effect of OCR-like Noise} 
We use Griko-Italian as our main evaluation pair due to the presence of its gold alignments, which can most accurately reflect the model's performance under a low-resource scenario.

We first benchmark model performance on clean and OCRed parallel text to quantify OCR-error effects on alignment (Table~\ref{table:noisyEffect}).
 We compute AER for the clean and OCRed versions of Griko-Italian by comparing their alignment against our manually created gold alignment. We benchmark five different models that lead to several observations.
 First, note that clean text always results in a better alignment for all models. Overall, \texttt{Giza++} performs best among the models, but note that it also suffers the largest drop in performance when faced with noisy text. On the other hand, a vanilla \texttt{awesome-align}, which is otherwise a state-of-the-art model for languages that were included in the pre-training of its underlying model, performs the worst, not being better than a simple~\texttt{IBM~1}. 
 
 We can thus conclude that OCR error does impact alignment quality for both statistical and neural based alignment models.~\label{baseline}
 
It is of note that for Griko-Italian every statistical model outperforms \texttt{awesome-align} in almost all cases. We hypothesize that this is due to the lack of structural knowledge; we deal with this in Section~\ref{bias}.~\texttt{awesome-align}'s low performance can also be explained by the fact that Griko is not well supported by its underlying representation model: Griko was not part of the pre-training language mix, and it does not use the same script as its closest language that was included in pre-training (Greek),\footnote{Modern Greek uses the Greek alphabet, while Griko uses the Latin alphabet.} an important factor according to~\citet{muller2021being}. Compared to statistical models, we also observe considerably fewer alignment pairs are produced by~\texttt{awesome-align} (Appendix~\ref{table:precisionAndRecall}), which might also be a contributing factor. 

\subsection{Making \texttt{awesome-align} Robust}
The performance of~\texttt{awesome-align} raises an intriguing question - Is the state-of-the-art neural based model capable to align noisy text, especially from low-resource languages. Given its general higher performance on many popular languages~\cite{dou-neubig-2021-word} and the stability between clean and noisy text,\footnote{Lowest AER difference between clean and noisy text amount to all models.} we make \texttt{awesome-align} as our main experiment target. 

\subsubsection{Low-Resource Setting}
We introduce structural bias and propose two models: model (\texttt{A}) and model (\texttt{B}) finetuned in unsupervised and supervised settings respectively. 

\begin{figure}
    \centering  
    \vspace{-1em}
    \includegraphics[width=7.5cm]{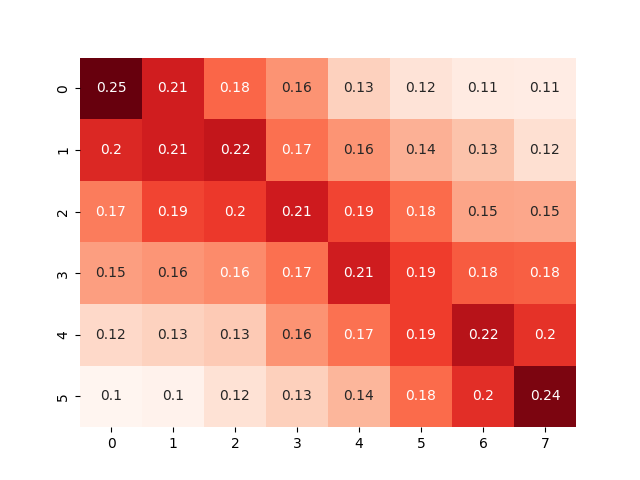}
    \caption{A sample 6 $\times$ 8 diagonal bias matrix. Darker color means stronger bias emphasis. We follow the same steps from~\citet{dyer2013simple} to calculate each position based on given rows and columns.}
    \label{fig:heatmap}
    \vspace{-1em}
\end{figure}

\paragraph{Structural Bias}
Structural alignment biases are widely used in statistical alignment models such as~\citet{brown1993mathematics, vogel1996hmm,och2003minimum,dyer2013simple}. However, it is a missing component in \texttt{awesome-align}. Following by~\citet{dyer2013simple}, we introduce diagonal bias and apply it on the top of \texttt{awesome-align}'s attention layer. We create \textit{(i)} a bias matrix $\mathit{M_{b}}$ based on the position of the alignment, where the positions near the diagonal of the alignment matrix have the higher weights (See Figure~\ref{fig:heatmap}); \textit{(ii)} a tune-able hyper-parameter $\lambda$ represents the weight of the bias. We set $\lambda$=1 for all low-resource language experiments; \textit{(iii)} an average matrix $\mathit{M_{avg}}$ that is the average of the original attention score, which is used for smoothing $\lambda$ to make it where 1 represents maximum bias and 0 means no bias at all. We update the original \texttt{awesome-align} attention score $\mathit{A_{sc}}$:~\label{bias}
\[
   \mathit{A_{sc}} = \lambda * \mathit{A_{sc}} +  (1-\lambda) * \mathit{A_{sc}} * \mathit{M_{b}} * \mathit{M_{avg}}
\]

\paragraph{Our proposed models}
For our unsupervised-finetuned model (\texttt{A}), we create the synthetically-noised data by introducing OCR-like noise on clean parallel data, and then simply finetune the baseline model with all available data from both clean and synthetic text.

\begin{table*}[t!]
\centering
\begin{tabular}{@{}l|cccc|ccc} 
 \toprule
 & \multicolumn{4}{c|}{English-French}& \multicolumn{3}{c}{English-German} \\
Test Set & Baseline & Unsup-FT  & Sup-FT & Reduction &  Baseline & Unsup-FT & Reduction \\
\midrule
\textsc{clean-clean}& 5.6 & \textbf{4.6} & 15.9 & 17.0\% & 17.9 & \textbf{15.2} & 15.1\%\\ 
\textsc{clean-synth}& 40.5 & 36.3 & \textbf{29.4} & 27.4\% & 43.8 & \textbf{39.4} & 10.0\%\\ 
\textsc{noisy-clean}& 39.2 & 34.2 & \textbf{28.6} & 27.0\% & 52.8 &\textbf{50} & 5.3\%\\ 
\textsc{noisy-noisy} & 53.6 & 46.1 & \textbf{37.3} & 30.4\% & 66.6 &\textbf{63.5} & 4.7\%\\ 
 \bottomrule
\end{tabular}
\caption{Result of \texttt{awesome-align} on English to French and German alignment. Both unsupervised and supervised finetuning with noise-induced data leads to big AER reduction when aligning noisy data. Reductions in German are less pronounced. Unsupervised finetuning with noisy data also improves clean-data alignment.}
\label{table:sumresults}
\end{table*}

\begin{table*}[t!]
\centering
\begin{tabular}{l | c c| c c| c c |c c} 
\toprule
&\multicolumn{2}{ c }{\texttt{IBM 1}} & \multicolumn{2}{| c }{\texttt{IBM 2}}& \multicolumn{2}{| c }{\texttt{Giza++}}& \multicolumn{2}{| c }{\texttt{fast-align}}\\

& Clean & OCRed & Clean & OCRed & Clean & OCRed & Clean & OCRed \\
\midrule
Baseline &  43.7 & 49.2& 37.3 & 43.4 & 14.5 & 20.8 & 19.8 & 25.7\\ 
Train w. clean& \textbf{40.2} & \textbf{45.7} & \textbf{32.7} & \textbf{38.1} & \textbf{13.1} & \textbf{19.0}  & \textbf{17.9}  & \textbf{24.2}\\ 
Train w. noise& 84.2 &  84.6 & 80.0 & 80.8  & 19.7 & 25.8 & 22.8 & 27.9\\ 
\bottomrule

\end{tabular}
\caption{Experiment on Griko-Italian, every statistical model benefits from training with additional clean data but suffers significant performance drops with synthetic noisy data, suggesting that traditional statistical models rely on clean text.}
\label{table:StatisticalModels}
\vspace{-1em}
\end{table*}

For the supervised-finetuned model (\texttt{B}), we first finetune an out-of-the-box \texttt{awesome-align} with the clean data from~\citet{anastasopoulos+al:2018:COLING} and~\citet{rijhwani-etal-2020-ocr} for Griko-Italian and Aiun-Japanese respectively, which  produces silver alignment. Next, we use the silver alignment as supervision to finetune \texttt{awesome-align} with synthetic noisy data.

We report the average plus-minus standard deviation of three runs for each model. Table~\ref{table:proposedModels} summarizes the results for our proposed models. We end up with around \textbf{\textit{50\%}} AER reduction for both endangered language pairs.

\subsubsection{High-Resource Setting}
We evaluate our data augmentation method on high-resource language pairs. Up to 400K synthetically noised English-French data was used for unsupervised finetuning. We also offer an additional reference data point, using 100K synthetic noised English-German data for unsupervised fine-tuning. 

For supervised finetuning, we use up to 1M synthetic data. As before, we use silver alignments from clean data as supervision to finetune its synthetic noisy version, which does not require any additional human annotation effort. 

Under both settings, model performance will plateau when adding more data. The results are summarized in Table~\ref{table:sumresults}. Both unsupervised and supervised finetunings with synthetically-noised data significantly improve alignment quality, especially for noisy test sets, in line with our previously presented results in low-resource settings.

\subsection{Addtional Data on Statistical Models}
We conduct additional experiments to find out whether training with additional data aids statistical models for endangered languages. We evaluate model performance on Griko-Italian.

We concatenate additional data to the examples comprising the test set. We first train the models with all 800 clean sentence pairs taken from ~\citet{rijhwani-etal-2020-ocr} (which include the 300 sentences of the test set). Next, instead of using clean data, we substitute it with synthetically noised data and train the models. 

The result is presented in Table~\ref{table:StatisticalModels}. For every statistical model, training with additional clean text reduces AER.
However, training with additional noisy text considerably hurts the models. The result shows that these statistical models rely on \textbf{\textit{clean text}} to improve, which is almost always \textbf{\textit{unavailable}} for endangered languages. This also implies that
investing time in manually cleaning OCR data could be effective for these models; however, it is not always possible and contradicts the goal of reducing the human effort in this work.

\section{Analysis and Discussion}
In this section we conduct several analyses to better understand our method. 

\paragraph{Incorporating Diagonal Bias}
As shown in Table~\ref{table:proposedModels},
our diagonal bias markedly improves every test case for both endangered language pairs. Note that the attention score will be increased significantly by adding bias, which will still be a valid input for the final alignment matrix due to its alignment extraction mechanism~\cite{dou-neubig-2021-word}. In this work, we only apply diagonal bias under low-resource settings since it was shown in~\citet{dou-neubig-2021-word} that growing heuristics such as grow-diag-final~\cite{koehn-etal-2005-edinburgh, och-ney-2000-improved} do not achieve promising results for multiple high-resource language test sets.

\begin{table}[tp]
\centering
\begin{tabular}{c c c} 
 \toprule
Test set & En-Fr & En-De  \\
\midrule
\textsc{clean-clean} & 5.6 & 17.9\\ 
\textsc{clean-noisy}& 40.5 & 43.8 \\ 
\textsc{noisy-clean}& 39.2 & 52.8 \\ 
\textsc{noisy-noisy}& 53.6 & 66.6 \\ 
\bottomrule

\end{tabular}
\caption{\texttt{awesome-align} baseline on En-Fr and En-De. OCR-like noise dramatically degrades the performance.}
\label{table:degradation}
\vspace{-1em}

\end{table}

\paragraph{Degradation of Alignment}
Table~\ref{table:degradation} presents the evaluation of four test sets for~\texttt{awesome-align} in English-French and English-German. We observe a significant decline in performance when OCR-like noise is introduced. For example, with clean parallel text, the AER for English-French is 5.6\%, but when OCR-like noise is added, the AER jumps to 53.6\%, almost a tenfold increase.

\paragraph{Size of synthetic data} We conduct quantitative analyses as shown in Figure~\ref{fig:sizeplot} to examine \texttt{awesome-align} with different sizes of English-French synthetic data under both unsupervised and supervised settings. For space economy reasons, here we only discuss the results of the more challenging noisy-noisy test set. Note that dramatic degradation of alignment is observed when applying OCR-like noise to clean text (see Table~\ref{table:degradation}). In general, the model produces better alignment as more data are used. However, there is also a trade-off on the clean-clean test set as its performance worsens in the supervised scenario. Keep in mind, though, that this situation is only observed in high-resource language pairs; for a low-resource language pair like our Griko-Italian, in limited ablation experiments we found that we have not reached the data saturation point yet as more data simply resulted in better performance for both clean and noisy text. 

\input{images/sizeplot}

\paragraph{Varying degrees of CER}
In a real-world scenario, the CER of OCRed data is typically unknown due to the absence of clean text. We investigate how different degrees of CER affect alignments by creating several English-French synthetic data with varying degrees of CER, testing them on \texttt{awesome-align}. We elaborate on the process and results in Appendix~\ref{veryCER}. The main finding is that higher CER leads to greater AER, which is expected. However, we also find that mixing with different degrees of CER generally produces better results than a fixed CER throughout the corpus, suggesting that our augmentation approach could also work on the unknown CER real-world scenario.

\paragraph{Statistical Model vs Neural Model}
The question of which model to use in practical scenarios, though, remains tricky to answer. Due to similarities between Griko and Italian and prolonged language contact over centuries, the two languages follow very similar syntax; as a result, their alignment is largely monotone, which benefits models like \texttt{Giza++} and \texttt{fast-align}. They outperform, in fact, the vanilla neural \texttt{awesome-align} model by a large margin (see Table~\ref{table:noisyEffect}). 
However, this will not always be the case. For example, most books with parallel data in the Archive of Indigenous Languages of Latin America (AILLA) mostly contain data between indigenous languages and one of Spanish or English. Now the two sides of the data come from different language families and a monotone alignment is not necessarily to be expected. In such cases, it could indeed be the case that a more adaptable neural model like \texttt{awesome-align}, aided by our data augmentation and diagonal biasing methods, could indeed be the best option.

\paragraph{Different side of OCR noise} An important insight derived from Table~\ref{table:sumresults} is that the performance of \texttt{awesome-align} deteriorates significantly more when both sides of the parallel data are noisy, as compared to when only one side is noisy. This is in fact encouraging for our envisioned application scenarios, since, as in the AILLA examples described above, we expect that OCRed parallel data in endangered languages will come with one side in a high-resource standardized language like English and Spanish which in turn we expect the OCR model to be able to adequately handle.\footnote{\citet{rijhwani-etal-2020-ocr} and \citet{rijhwani2021lexically} make similar observations on all endangered language datasets they work with.}

\section{Related Work}
Our work is a natural extension of previous word alignment work. A robust alignment tool for low-resource languages benefits MT systems~\cite{xiang2010diversify,levinboim-chiang-2015-multi,beloucif-etal-2016-improving,nagata-etal-2020-supervised}, or speech recognition~\cite{anastasopoulos2018leveraging}, especially if sentence-level alignment tools like LASER~\cite{artetxe2019massively,chaudhary2019low} do not cover all languages, so one may need to fall-back to word-level alignment heuristics to inform sentence-alignment models like \texttt{Hunalign}~\cite{varga2007parallel}. 

Research on word-level alignment started with statistical models, with the IBM Translation Models~\cite{brown1993mathematics} serving as the foundation for many popular statistical word aligners~\cite{och-ney-2000-improved,och2003systematic,och2003minimum,tiedemann-etal-2016-phrase,vogel1996hmm,och2003minimum,gao-vogel-2008-parallel,dyer2013simple}. In recent years, different neural network based alignment models gained in popularity including end-to-end based~\cite{zenkel-etal-2020-end,wu2022mirroralign,chen-etal-2021-mask}, MT-based~\cite{chen-etal-2020-accurate}, and pre-training based~\cite{garg2019jointly,dou-neubig-2021-word}. As~\texttt{awesome-align} achieves the overall highest performance, we choose to focus on~\texttt{awesome-align} in this work. 

Some works involve improving word-level alignment for low-resource languages such as utilizing semantic information~\cite{beloucif2016improving,pourdamghani-etal-2018-using}, multi-task learning~\cite{levinboim-chiang-2015-multi}, and combining complementary word alignments~\cite{xiang-etal-2010-diversify}. None of the previous work, though, to our knowledge, tackles the problem of aligning data with OCR-like noise on one or both sides. The idea of augmenting training data is not new and has been applied in many areas and applications.~\citet{marton-etal-2009-improved} augment data with paraphrases taken from other languages to improve low-resource language alignments. While potentially orthogonal to our approach, this idea is largely inapplicable to our endangered language settings, as we often have to work with the only available datasets for these particular languages. Applying structure alignment bias on statistical and neural models is also a well-studied area~\cite{cohn-etal-2016-incorporating,brown1993mathematics, vogel1996hmm,och2003minimum,dyer2013simple}. However, to the best of our knowledge, we are the first to apply it to low-resource languages, proving that such an approach can greatly aid the real endangered language data.

\section{Conclusion}
In this work, we benchmark several popular word alignment models under OCR noisy settings with high- and low-resource language pairs, conducting several studies to investigate the relationship between OCR noise and alignment quality. We propose a simple yet effective approach to create realistic OCR-like synthetic data and make the state-of-the-art neural \texttt{awesome-align} model more robust by leveraging structural bias. 
Our work paves the way for future word-level alignment-related research on underrepresented languages. As part of this paper, we also release a total of 4,101 ground truth word alignment data for Griko-Italian, which can be a useful resource to investigate word- and sentence-level alignment techniques on practical endangered language scenarios. 

\section{Limitations}
Using AER as the main evaluation metric could be a limitation of our work as it might be misleading in some cases~\cite{fraser2007measuring}. Another limitation, of course, is that we only manage to explore the tip of the iceberg given the sheer number of endangered languages. While we are confident in the results of both low-resource language pairs, our experiments on Ainu-Japenese could potentially lead to inaccurate AER since we use the automatically generated silver alignment. In the future, we hope to eventually annotate it with either the help of native speakers or dictionaries. We also plan to explore other alternative metrics and expand our alignment benchmark on as many endangered languages as possible. 

\section*{Acknowledgements}
We are thankful to Shruti Rijhwani and Graham Neubig, as well as the anonymous reviewers, for their valuable comments on the early stages of this work. We would also like to thank Sina Ahmadi for his useful feedback and the GMU Office of Research Computing for the computing resources. This work was supported by NEH Award PR-276810-21 as well as through a GMU OSCAR award for undergraduate research.

\bibliography{custom.bib}
\bibliographystyle{acl_natbib}

\clearpage
\newpage
\appendix

\section{Evaluation metric}
We calculate precision, recall and alignment error rate as described in~\citet{och2003systematic}, where $A$ is a set of alignments to compare, $S$ is a set of gold alignments, and $P$ is the union of $A$ and possible alignments in $S$. We then compute AER with:\\~\label{metric}

\begin{tabular}{cc}
    $\text{Precision} = \frac{| A \cap P |}{| A |} $ & $\text{Recall} = \frac{| A \cap S |}{| S |}$ 
\end{tabular}
\[
\text{AER} (S, P; A) = 1 - \frac{| A \cap S + A \cap P |}{| A + S |}
\]

\section{Additional Analyses}

\subsection{Varying degrees of CER}
We create eight 100k English-French synthetic datasets with different CER for each: ``unified'' datasets with exactly the same CER on both sides: 2, 5, 10, and one with mixed CER with equally shared portions with 2, 5 and 10 CER; and ``varying'' datasets with slightly different CER between the English and French side, but in the same range as the others. 
We then finetune the model with these synthetic training datasets and compare against the ``skyline'' result presented before, where the augmentation matched the level of the true CER.

The results are presented in Table~\ref{table:VaryingCER} for both the unsupervised- and the supervised-finetuning setting. Encouragingly, despite different CER in the augmentation data, there are no significant performance differences in most cases, especially for the unsupervised setting. 
Of course, levels of noise that match the true level tend to perform better or close to best overall. On the other hand, high levels of noise that lead to very high word error rate (WER)\footnote{For example, a CER of around 10 translates to a WER of more than 70, meaning that (approximately) only 3 out of 10 words are correct.} cause a large degradation in the performance of the supervised finetuning approach, but do not seem to significantly affect the unsupervised approach.

\begin{table*}[tp]
\centering
\begin{tabular}{@{}r|c@{ }c|c@{ }c|c@{ }c|c@{ }c|c@{ }c} 
\toprule
&\multicolumn{2}{c|}{\texttt{IBM 1}} & \multicolumn{2}{c|}{\texttt{IBM 2}}& \multicolumn{2}{c|}{\texttt{Giza++}}& \multicolumn{2}{c|}{\texttt{fast-align}}&\multicolumn{2}{c}{\texttt{awesome}} \\

& Clean & OCR & Clean & OCR & Clean & OCR & Clean & OCR & Clean & OCR \\
\midrule

\# of pairs & 3844 & 3839 & 3833 & 3855 & 3810 & 3813 & 3801 & 3794 & 2978 & 2969\\ 
\midrule

Precision& 58.2 & 52.5 & 64.9 & 58.6 & 88.7 & 82.2 & 83.4 & 77.3 & 65.3 & 64 \\ 
Recall& 54.5 & 49.2 & 60.7 & 54.8 & 82.4 & 76.4 & 77.3 & 71.5 & 47.4 & 44.2\\ 

\bottomrule

\end{tabular}
\caption{Comparing the number of alignment pairs produced by models on Griko-Italian. \texttt{awesome-align} produces almost 25\% less alignment pairs, resulting in markedly lower precision/recall and higher AER.}
\label{table:precisionAndRecall}
\end{table*}

Even more encouragingly, an augmented dataset that uses a mixture of different target CER (such as having a third of the dataset having a CER around 2, a third with CER around 5, and a third around 10 -- named ``mixed'' in Table~\ref{table:VaryingCER}) in the supervised setting further outperforms the \textit{informed skyline} which uses additional knowledge that might not be available (the true CER of the data to be aligned). For instance, in the clean-noisy test set this model reduces AER by a further 5\% (from 31.1 to 29.3) and on the clean-clean test set it reduces AER by 19\% (from 6.8 to 5.5).
This means that our augmentation approach with varying levels of noise could be applied to any scenario, even if one does not know the level of noise present in the data-to-be-aligned.~\label{veryCER}

\begin{table*}[t!]
\centering
\begin{tabular}{ll@{ }c@{ }c|c@{ }c|c@{ }c|c@{ }c@{ }} 
 \toprule
 \multicolumn{2}{c}{CER (WER) on}& \multicolumn{2}{c|}{Clean-Clean} & \multicolumn{2}{c|}{Clean-Noisy} & \multicolumn{2}{c|}{Noisy-Clean}& \multicolumn{2}{c}{Noisy-Noisy}\\

 \multicolumn{2}{c}{Synthetic Data}& \small \textsc{unsup-ft} & \small \textsc{sup-ft} & \small \textsc{unsup-ft} & \small \textsc{sup-ft} & \small \textsc{unsup-ft} & \small \textsc{sup-ft} & \small \textsc{unsup-ft} & \small \textsc{sup-ft}  \\
\midrule
\multicolumn{8}{l}{Skyline: Using exactly the CER of the test set}\\
& 7.4-4.8 (59.7-51.7) & 4.3 &6.8 & 37 & 31.1& 34.4 & 30 & 46.9  & 40.3  \\
\midrule
\multicolumn{8}{l}{Unified: Exactly the same CER on both sides}\\
&2-2 (32.2-29.1) & 4.1 & \textbf{5.1} & 37.5 & 33.3 & 35.1 & 30.1 & 48.4 & 41.8\\
&5-5 (55.1-52.4) & 4.3 & 7.4 & 37.2 & 30.4 & 34.6 & 29.8 & 47.4 & \textbf{40.0}\\
&10-10 (72.2-71) & 4.5 & 32.8 & \textbf{36.8} & 36.2 & 34.5 & 47.7 & 46.8 & 47.3\\
&mixed (55-54.1) & 5.4 & 5.5 & 37.3 & 39.6 & 35.3 & 39.8 & 47.7  & 54.2\\ 
\midrule

\multicolumn{8}{l}{Varying CER between the two parallel sides}\\
&1.6-2.1 (27.8-29.6) & \textbf{4.0} & 6.2 & 37.6 & 31.6 & 35 & 30.6 & 48.4 & 41.9\\
&4.1-5.1 (49.6-52.5) & 4.3 & 7.7 & 37& 29.8 & 34.7& 30.1 & 47.2 & 40.2\\
&8.1-9.6 (66.9-69.5) & 4.5 & 31.3 & 37.1& 36.2 & \textbf{34.3} & 46.3 & \textbf{46.7} & 46.9\\
&mixed (47.9-50) & 4.2& 5.6 & 37 & \textbf{29.3}& 34.6& \textbf{29.7} & 47.2 & 40.4\\
\bottomrule

\end{tabular}
\caption{AER comparison for varying CER in 100K English-French augmented data used for either unsupervised or supervised finetuning. We highlight the best result under each setting and test set. Overall, most models' performance is close to the baseline, but varying amounts of noise (mixed) lead to generally the best results. Too high amounts of noise (e.g. CER around 10 with WER approaching 70) hurts the supervised approach.}
\label{table:VaryingCER}
\end{table*}

\end{document}

%% file: images/sizeplot.tex
\pgfplotstableread[row sep=\\,col sep=&]{
num & cc & cs & sc & ss \\
0.8& 5.6&40.5& 39.2&  53.6 \\
1.5 & 4.9&  39.7&37.8& 52.6\\
3 & 4.8&  39.2&37.3& 51.9 \\
12 &4.4&  38.3&36.1& 49.4\\
 50 & 4.2& 37.3&  34.8& 47.7\\
 100 & 4.3&  37& 34.4& 46.9\\
 200 &4.5& 36.7& 34.1& 46.5\\
400& 4.6& 36.3& 34.2& 46.1\\
}\unsupfrdata
\pgfplotstableread[row sep=\\,col sep=&]{
num & cc & cs & sc & ss \\
0.8& 5.6&40.5& 39.2&  53.6 \\
1.5 & 6.1& 38.2& 36.2& 47.7\\
6 &5.8&  37.6&35.9& 47.3\\
25 &6.4&  36.8& 35.4& 46.5\\
100 &6.8& 31.1& 30& 40.3\\
400& 12.5& 29.2 & 28.6&  37.9\\
1000& 15.9& 29.4& 28.6& 37.3\\
}\supfrdata

\pgfplotsset{roy plot/.style={
            every axis plot post/.style={/pgf/number format/fixed},
            width=7.8cm,
            height=4cm,
            ymajorgrids=false,
            yminorgrids=false,
            every x tick label/.append style={font=\tiny},
            every y tick label/.append style={font=\tiny},
            ymax=55,ymin=0,
            tick pos=left,
            axis y line*=left,
            axis x line*=bottom,
            nodes near coords,
            every node near coord/.append style={font=\tiny,color=black},
            enlarge x limits=0.1,
            title style={yshift=-.1cm,font=\small},
            ylabel near ticks,
            xlabel near ticks,
    }
}

\begin{figure*}[t]
    \centering
\begin{tabular}{cc}
    \begin{tikzpicture}[trim left=5.5cm,trim right=-0.3cm]
    \begin{semilogxaxis}[roy plot,
    title={Unsupervised-Finetuning},
    log ticks with fixed point, ylabel={AER},xlabel={Additional Synthetic Data (K)},
    legend style={at={(0.97,0.6)},anchor=south west,font=\tiny,},
    ]
    \addlegendimage{empty legend};
    \addlegendentry{\hspace{-.6cm}\textbf{Test Set}};
    \addlegendentry{synth-synth};
    \addplot [mark=diamond] table[x=num,y=ss]{\unsupfrdata};
    \addplot [mark=square*] table[x=num,y=cs]{\unsupfrdata};
    \addlegendentry{clean-synth};
    \addplot [mark=o, every node near coord/.append style={yshift=-12pt,anchor=south}] table[x=num,y=sc]{\unsupfrdata};
    \addlegendentry{synth-clean};
    \addplot [mark=x] table[x=num,y=cc]{\unsupfrdata};
    \addlegendentry{clean-clean};
    \end{semilogxaxis}
\end{tikzpicture} & 
    \begin{tikzpicture}[trim left=-8.5cm,trim right=0cm]
    \begin{semilogxaxis}[roy plot,
    title={Supervised-Finetuning},
    log ticks with fixed point, xlabel={Additional Synthetic Data (K)},
    ]
    \addplot [mark=x] table[x=num,y=cc]{\supfrdata};
    \addplot [mark=diamond] table[x=num,y=ss]{\supfrdata};
    \addplot [mark=square*] table[x=num,y=cs]{\supfrdata};
    \addplot [mark=o, every node near coord/.append style={yshift=-12pt,anchor=south}] table[x=num,y=sc]{\supfrdata};
    \end{semilogxaxis}
\end{tikzpicture}
\end{tabular}
    \caption{Ablation on English-French with varying degrees of additional synthetically noised data. Notice the log scale on the x-axis. The left-most point corresponds to no additional synthetic data (baseline). More data reduce AER for noisy test sets, especially in the supervised finetuning setting.}
    \label{fig:sizeplot}
    \vspace{1em}
\end{figure*}

%% file: eacl2023.bbl
\begin{thebibliography}{44}
\expandafter\ifx\csname natexlab\endcsname\relax\def\natexlab#1{#1}\fi

\bibitem[{Alpert-Abrams(2016)}]{alpert2016machine}
Hannah Alpert-Abrams. 2016.
\newblock Machine reading the primeros libros.
\newblock \emph{Digital Humanities Quarterly}, 10(4).

\bibitem[{Amrhein and Clematide(2018)}]{amrhein2018supervised}
Chantal Amrhein and Simon Clematide. 2018.
\newblock Supervised ocr error detection and correction using statistical and
  neural machine translation methods.
\newblock \emph{Journal for Language Technology and Computational Linguistics
  (JLCL)}, 33(1):49--76.

\bibitem[{Anastasopoulos and Chiang(2018)}]{anastasopoulos2018leveraging}
Antonios Anastasopoulos and David Chiang. 2018.
\newblock Leveraging translations for speech transcription in low-resource
  settings.
\newblock \emph{Proc. Interspeech 2018}, pages 1279--1283.

\bibitem[{Anastasopoulos et~al.(2018)Anastasopoulos, Lekakou, Quer, Zimianiti,
  DeBenedetto, and Chiang}]{anastasopoulos+al:2018:COLING}
Antonios Anastasopoulos, Marika Lekakou, Josep Quer, Eleni Zimianiti, Justin
  DeBenedetto, and David Chiang. 2018.
\newblock Part-of-speech tagging on an endangered language: a parallel
  griko-italian resource.
\newblock In \emph{Proceedings of the 27th International Conference on
  Computational Linguistics}, pages 2529--2539.

\bibitem[{Artetxe and Schwenk(2019)}]{artetxe2019massively}
Mikel Artetxe and Holger Schwenk. 2019.
\newblock Massively multilingual sentence embeddings for zero-shot
  cross-lingual transfer and beyond.
\newblock \emph{Transactions of the Association for Computational Linguistics},
  7:597--610.

\bibitem[{Beloucif et~al.(2016{\natexlab{a}})Beloucif, Saers, and
  Wu}]{beloucif-etal-2016-improving}
Meriem Beloucif, Markus Saers, and Dekai Wu. 2016{\natexlab{a}}.
\newblock \href {https://aclanthology.org/W16-4507} {Improving word alignment
  for low resource languages using {E}nglish monolingual {SRL}}.
\newblock In \emph{Proceedings of the Sixth Workshop on Hybrid Approaches to
  Translation ({H}y{T}ra6)}, pages 51--60, Osaka, Japan. The COLING 2016
  Organizing Committee.

\bibitem[{Beloucif et~al.(2016{\natexlab{b}})Beloucif, Saers, and
  Wu}]{beloucif2016improving}
Meriem Beloucif, Markus Saers, and Dekai Wu. 2016{\natexlab{b}}.
\newblock Improving word alignment for low resource languages using english
  monolingual srl.
\newblock In \emph{Proceedings of the Sixth Workshop on Hybrid Approaches to
  Translation (HyTra6)}, pages 51--60.

\bibitem[{Bird(2020)}]{bird-2020-sparse}
Steven Bird. 2020.
\newblock \href {https://doi.org/10.1162/coli_a_00387} {Sparse transcription}.
\newblock \emph{Computational Linguistics}, 46(4):713--744.

\bibitem[{Brown et~al.(1993)Brown, Della~Pietra, Della~Pietra, and
  Mercer}]{brown1993mathematics}
Peter~F Brown, Stephen~A Della~Pietra, Vincent~J Della~Pietra, and Robert~L
  Mercer. 1993.
\newblock The mathematics of statistical machine translation: Parameter
  estimation.
\newblock \emph{Computational linguistics}, 19(2):263--311.

\bibitem[{Bustamante et~al.(2020)Bustamante, Oncevay, and
  Zariquiey}]{bustamante2020no}
Gina Bustamante, Arturo Oncevay, and Roberto Zariquiey. 2020.
\newblock No data to crawl? monolingual corpus creation from pdf files of truly
  low-resource languages in peru.
\newblock In \emph{Proceedings of the 12th Language Resources and Evaluation
  Conference}, pages 2914--2923.

\bibitem[{Chaudhary et~al.(2019)Chaudhary, Tang, Guzm{\'a}n, Schwenk, and
  Koehn}]{chaudhary2019low}
Vishrav Chaudhary, Yuqing Tang, Francisco Guzm{\'a}n, Holger Schwenk, and
  Philipp Koehn. 2019.
\newblock Low-resource corpus filtering using multilingual sentence embeddings.
\newblock \emph{WMT 2019}, page 263.

\bibitem[{Chen et~al.(2021)Chen, Sun, and Liu}]{chen-etal-2021-mask}
Chi Chen, Maosong Sun, and Yang Liu. 2021.
\newblock \href {https://doi.org/10.18653/v1/2021.acl-long.369} {Mask-align:
  Self-supervised neural word alignment}.
\newblock In \emph{Proceedings of the 59th Annual Meeting of the Association
  for Computational Linguistics and the 11th International Joint Conference on
  Natural Language Processing (Volume 1: Long Papers)}, pages 4781--4791,
  Online. Association for Computational Linguistics.

\bibitem[{Chen et~al.(2020)Chen, Liu, Chen, Jiang, and
  Liu}]{chen-etal-2020-accurate}
Yun Chen, Yang Liu, Guanhua Chen, Xin Jiang, and Qun Liu. 2020.
\newblock \href {https://doi.org/10.18653/v1/2020.emnlp-main.42} {Accurate word
  alignment induction from neural machine translation}.
\newblock In \emph{Proceedings of the 2020 Conference on Empirical Methods in
  Natural Language Processing (EMNLP)}, pages 566--576, Online. Association for
  Computational Linguistics.

\bibitem[{Cohn et~al.(2016)Cohn, Hoang, Vymolova, Yao, Dyer, and
  Haffari}]{cohn-etal-2016-incorporating}
Trevor Cohn, Cong Duy~Vu Hoang, Ekaterina Vymolova, Kaisheng Yao, Chris Dyer,
  and Gholamreza Haffari. 2016.
\newblock \href {https://doi.org/10.18653/v1/N16-1102} {Incorporating
  structural alignment biases into an attentional neural translation model}.
\newblock In \emph{Proceedings of the 2016 Conference of the North {A}merican
  Chapter of the Association for Computational Linguistics: Human Language
  Technologies}, pages 876--885, San Diego, California. Association for
  Computational Linguistics.

\bibitem[{Devlin et~al.(2019)Devlin, Chang, Lee, and
  Toutanova}]{devlin2018bert}
Jacob Devlin, Ming-Wei Chang, Kenton Lee, and Kristina Toutanova. 2019.
\newblock Bert: Pre-training of deep bidirectional transformers for language
  understanding.
\newblock In \emph{Proceedings of the 2019 Conference of the North American
  Chapter of the Association for Computational Linguistics: Human Language
  Technologies, Volume 1 (Long and Short Papers)}, pages 4171--4186.

\bibitem[{Dou and Neubig(2021)}]{dou-neubig-2021-word}
Zi-Yi Dou and Graham Neubig. 2021.
\newblock \href {https://doi.org/10.18653/v1/2021.eacl-main.181} {Word
  alignment by fine-tuning embeddings on parallel corpora}.
\newblock In \emph{Proceedings of the 16th Conference of the European Chapter
  of the Association for Computational Linguistics: Main Volume}, pages
  2112--2128, Online. Association for Computational Linguistics.

\bibitem[{Dyer et~al.(2013)Dyer, Chahuneau, and Smith}]{dyer2013simple}
Chris Dyer, Victor Chahuneau, and Noah~A Smith. 2013.
\newblock A simple, fast, and effective reparameterization of ibm model 2.
\newblock In \emph{Proceedings of the 2013 Conference of the North American
  Chapter of the Association for Computational Linguistics: Human Language
  Technologies}, pages 644--648.

\bibitem[{Fraser and Marcu(2007)}]{fraser2007measuring}
Alexander Fraser and Daniel Marcu. 2007.
\newblock Measuring word alignment quality for statistical machine translation.
\newblock \emph{Computational Linguistics}, 33(3):293--303.

\bibitem[{Gao and Vogel(2008)}]{gao-vogel-2008-parallel}
Qin Gao and Stephan Vogel. 2008.
\newblock \href {https://aclanthology.org/W08-0509} {Parallel implementations
  of word alignment tool}.
\newblock In \emph{Software Engineering, Testing, and Quality Assurance for
  Natural Language Processing}, pages 49--57, Columbus, Ohio. Association for
  Computational Linguistics.

\bibitem[{Garg et~al.(2019)Garg, Peitz, Nallasamy, and
  Paulik}]{garg2019jointly}
Sarthak Garg, Stephan Peitz, Udhyakumar Nallasamy, and Matthias Paulik. 2019.
\newblock \href {https://doi.org/10.18653/v1/D19-1453} {Jointly learning to
  align and translate with transformer models}.
\newblock In \emph{Proceedings of the 2019 Conference on Empirical Methods in
  Natural Language Processing and the 9th International Joint Conference on
  Natural Language Processing (EMNLP-IJCNLP)}, pages 4453--4462, Hong Kong,
  China. Association for Computational Linguistics.

\bibitem[{Ignat et~al.(2022)Ignat, Maillard, Chaudhary, and
  Guzm{\'a}n}]{ignat2022ocr}
Oana Ignat, Jean Maillard, Vishrav Chaudhary, and Francisco Guzm{\'a}n. 2022.
\newblock {OCR} {I}mproves {M}achine {T}ranslation for {L}ow-{R}esource
  {L}anguages.
\newblock In \emph{Findings of the Association for Computational Linguistics:
  ACL 2022}, pages 1164--1174.

\bibitem[{Koehn(2005)}]{koehn2005europarl}
Philipp Koehn. 2005.
\newblock Europarl: A parallel corpus for statistical machine translation.
\newblock In \emph{Proceedings of machine translation summit x: papers}, pages
  79--86.

\bibitem[{Koehn et~al.(2005)Koehn, Axelrod, Birch~Mayne, Callison-Burch,
  Osborne, and Talbot}]{koehn-etal-2005-edinburgh}
Philipp Koehn, Amittai Axelrod, Alexandra Birch~Mayne, Chris Callison-Burch,
  Miles Osborne, and David Talbot. 2005.
\newblock \href {https://aclanthology.org/2005.iwslt-1.8} {{E}dinburgh system
  description for the 2005 {IWSLT} speech translation evaluation}.
\newblock In \emph{Proceedings of the Second International Workshop on Spoken
  Language Translation}, Pittsburgh, Pennsylvania, USA.

\bibitem[{Levinboim and Chiang(2015)}]{levinboim-chiang-2015-multi}
Tomer Levinboim and David Chiang. 2015.
\newblock \href {https://doi.org/10.3115/v1/N15-1129} {Multi-task word
  alignment triangulation for low-resource languages}.
\newblock In \emph{Proceedings of the 2015 Conference of the North {A}merican
  Chapter of the Association for Computational Linguistics: Human Language
  Technologies}, pages 1221--1226, Denver, Colorado. Association for
  Computational Linguistics.

\bibitem[{Marton et~al.(2009)Marton, Callison-Burch, and
  Resnik}]{marton-etal-2009-improved}
Yuval Marton, Chris Callison-Burch, and Philip Resnik. 2009.
\newblock \href {https://aclanthology.org/D09-1040} {Improved statistical
  machine translation using monolingually-derived paraphrases}.
\newblock In \emph{Proceedings of the 2009 Conference on Empirical Methods in
  Natural Language Processing}, pages 381--390, Singapore. Association for
  Computational Linguistics.

\bibitem[{Mihalcea and Pedersen(2003)}]{mihalcea2003evaluation}
Rada Mihalcea and Ted Pedersen. 2003.
\newblock An evaluation exercise for word alignment.
\newblock In \emph{Proceedings of the HLT-NAACL 2003 Workshop on Building and
  using parallel texts: data driven machine translation and beyond}, pages
  1--10.

\bibitem[{Muller et~al.(2021)Muller, Anastasopoulos, Sagot, and
  Seddah}]{muller2021being}
Benjamin Muller, Antonios Anastasopoulos, Beno{\^\i}t Sagot, and Djam{\'e}
  Seddah. 2021.
\newblock {W}hen {B}eing {U}nseen from {mBERT} is just the {B}eginning:
  {H}andling {N}ew {L}anguages with {M}ultilingual {L}anguage {M}odels.
\newblock In \emph{Proceedings of the 2021 Conference of the North American
  Chapter of the Association for Computational Linguistics: Human Language
  Technologies}, pages 448--462.

\bibitem[{Nagata et~al.(2020)Nagata, Chousa, and
  Nishino}]{nagata-etal-2020-supervised}
Masaaki Nagata, Katsuki Chousa, and Masaaki Nishino. 2020.
\newblock \href {https://doi.org/10.18653/v1/2020.emnlp-main.41} {A supervised
  word alignment method based on cross-language span prediction using
  multilingual {BERT}}.
\newblock In \emph{Proceedings of the 2020 Conference on Empirical Methods in
  Natural Language Processing (EMNLP)}, pages 555--565, Online. Association for
  Computational Linguistics.

\bibitem[{Och(2003)}]{och2003minimum}
Franz~Josef Och. 2003.
\newblock Minimum error rate training in statistical machine translation.
\newblock In \emph{Proceedings of the 41st annual meeting of the Association
  for Computational Linguistics}, pages 160--167.

\bibitem[{Och and Ney(2000)}]{och-ney-2000-improved}
Franz~Josef Och and Hermann Ney. 2000.
\newblock \href {https://doi.org/10.3115/1075218.1075274} {Improved statistical
  alignment models}.
\newblock In \emph{Proceedings of the 38th Annual Meeting of the Association
  for Computational Linguistics}, pages 440--447, Hong Kong. Association for
  Computational Linguistics.

\bibitem[{Och and Ney(2003)}]{och2003systematic}
Franz~Josef Och and Hermann Ney. 2003.
\newblock A systematic comparison of various statistical alignment models.
\newblock \emph{Computational linguistics}, 29(1):19--51.

\bibitem[{Pourdamghani et~al.(2018)Pourdamghani, Ghazvininejad, and
  Knight}]{pourdamghani-etal-2018-using}
Nima Pourdamghani, Marjan Ghazvininejad, and Kevin Knight. 2018.
\newblock \href {https://doi.org/10.18653/v1/N18-2083} {Using word vectors to
  improve word alignments for low resource machine translation}.
\newblock In \emph{Proceedings of the 2018 Conference of the North {A}merican
  Chapter of the Association for Computational Linguistics: Human Language
  Technologies, Volume 2 (Short Papers)}, pages 524--528, New Orleans,
  Louisiana. Association for Computational Linguistics.

\bibitem[{Rigaud et~al.(2019)Rigaud, Doucet, Coustaty, and
  Moreux}]{rigaud2019icdar}
Christophe Rigaud, Antoine Doucet, Micka{\"e}l Coustaty, and Jean-Philippe
  Moreux. 2019.
\newblock {ICDAR 2019 competition on post-OCR text correction}.
\newblock In \emph{2019 international conference on document analysis and
  recognition (ICDAR)}, pages 1588--1593. IEEE.

\bibitem[{Rijhwani et~al.(2020)Rijhwani, Anastasopoulos, and
  Neubig}]{rijhwani-etal-2020-ocr}
Shruti Rijhwani, Antonios Anastasopoulos, and Graham Neubig. 2020.
\newblock \href {https://doi.org/10.18653/v1/2020.emnlp-main.478} {{OCR} {P}ost
  {C}orrection for {E}ndangered {L}anguage {T}exts}.
\newblock In \emph{Proceedings of the 2020 Conference on Empirical Methods in
  Natural Language Processing (EMNLP)}, pages 5931--5942, Online. Association
  for Computational Linguistics.

\bibitem[{Rijhwani et~al.(2021)Rijhwani, Rosenblum, Anastasopoulos, and
  Neubig}]{rijhwani2021lexically}
Shruti Rijhwani, Daisy Rosenblum, Antonios Anastasopoulos, and Graham Neubig.
  2021.
\newblock Lexically aware semi-supervised learning for ocr post-correction.
\newblock \emph{Transactions of the Association for Computational Linguistics},
  9:1285--1302.

\bibitem[{Tiedemann et~al.(2016)Tiedemann, Cap, Kanerva, Ginter, Stymne,
  {\"O}stling, and Weller-Di~Marco}]{tiedemann-etal-2016-phrase}
J{\"o}rg Tiedemann, Fabienne Cap, Jenna Kanerva, Filip Ginter, Sara Stymne,
  Robert {\"O}stling, and Marion Weller-Di~Marco. 2016.
\newblock \href {https://doi.org/10.18653/v1/W16-2326} {Phrase-based {SMT} for
  {F}innish with more data, better models and alternative alignment and
  translation tools}.
\newblock In \emph{Proceedings of the First Conference on Machine Translation:
  Volume 2, Shared Task Papers}, pages 391--398, Berlin, Germany. Association
  for Computational Linguistics.

\bibitem[{Van~Strien et~al.(2020)Van~Strien, Beelen, Ardanuy, Hosseini,
  McGillivray, and Colavizza}]{van2020assessing}
Daniel Van~Strien, Kaspar Beelen, Mariona~Coll Ardanuy, Kasra Hosseini, Barbara
  McGillivray, and Giovanni Colavizza. 2020.
\newblock {Assessing the impact of OCR quality on downstream NLP tasks}.
\newblock \emph{SCITEPRESS-Science and Technology Publications}.

\bibitem[{Varga et~al.(2007)Varga, Hal{\'a}csy, Kornai, Nagy, N{\'e}meth, and
  Tr{\'o}n}]{varga2007parallel}
D{\'a}niel Varga, P{\'e}ter Hal{\'a}csy, Andr{\'a}s Kornai, Viktor Nagy,
  L{\'a}szl{\'o} N{\'e}meth, and Viktor Tr{\'o}n. 2007.
\newblock Parallel corpora for medium density languages.
\newblock \emph{Amsterdam Studies In The Theory And History Of Linguistic
  Science Series 4}, 292:247.

\bibitem[{Vilar et~al.(2006)Vilar, Popovi{\'c}, and Ney}]{vilar2006aer}
David Vilar, Maja Popovi{\'c}, and Hermann Ney. 2006.
\newblock {AER: Do we need to “improve” our alignments?}
\newblock In \emph{Proceedings of the Third International Workshop on Spoken
  Language Translation: Papers}.

\bibitem[{Vogel et~al.(1996)Vogel, Ney, and Tillmann}]{vogel1996hmm}
Stephan Vogel, Hermann Ney, and Christoph Tillmann. 1996.
\newblock Hmm-based word alignment in statistical translation.
\newblock In \emph{COLING 1996 Volume 2: The 16th International Conference on
  Computational Linguistics}.

\bibitem[{Wu et~al.(2022)Wu, Ding, Yang, and Li}]{wu2022mirroralign}
Di~Wu, Liang Ding, Shuo Yang, and Mingyang Li. 2022.
\newblock Mirroralign: A super lightweight unsupervised word alignment model
  via cross-lingual contrastive learning.
\newblock In \emph{Proceedings of the 19th International Conference on Spoken
  Language Translation (IWSLT 2022)}, pages 83--91.

\bibitem[{Xiang et~al.(2010{\natexlab{a}})Xiang, Deng, and
  Zhou}]{xiang2010diversify}
Bing Xiang, Yonggang Deng, and Bowen Zhou. 2010{\natexlab{a}}.
\newblock Diversify and combine: Improving word alignment for machine
  translation on low-resource languages.
\newblock In \emph{Proceedings of the ACL 2010 Conference Short Papers}, pages
  22--26.

\bibitem[{Xiang et~al.(2010{\natexlab{b}})Xiang, Deng, and
  Zhou}]{xiang-etal-2010-diversify}
Bing Xiang, Yonggang Deng, and Bowen Zhou. 2010{\natexlab{b}}.
\newblock \href {https://aclanthology.org/P10-2005} {Diversify and combine:
  Improving word alignment for machine translation on low-resource languages}.
\newblock In \emph{Proceedings of the {ACL} 2010 Conference Short Papers},
  pages 22--26, Uppsala, Sweden. Association for Computational Linguistics.

\bibitem[{Zenkel et~al.(2020)Zenkel, Wuebker, and
  DeNero}]{zenkel-etal-2020-end}
Thomas Zenkel, Joern Wuebker, and John DeNero. 2020.
\newblock \href {https://doi.org/10.18653/v1/2020.acl-main.146} {End-to-end
  neural word alignment outperforms {GIZA}++}.
\newblock In \emph{Proceedings of the 58th Annual Meeting of the Association
  for Computational Linguistics}, pages 1605--1617, Online. Association for
  Computational Linguistics.

\end{thebibliography}
